\documentclass[letterpaper, 10 pt, conference]{ieeeconf}  

\IEEEoverridecommandlockouts                              

\usepackage{cite}
\usepackage{amsmath,amssymb,amsfonts}
\usepackage{algorithmic}
\usepackage{graphicx}
\usepackage{textcomp}
\usepackage{xcolor}
\usepackage{algorithm}
\usepackage{algorithmic}
\usepackage[hyphens]{url}
\usepackage[colorlinks,urlcolor=blue,linkcolor=blue,citecolor=blue]{hyperref}
\def\BibTeX{{\rm B\kern-.05em{\sc i\kern-.025em b}\kern-.08em
    T\kern-.1667em\lower.7ex\hbox{E}\kern-.125emX}}
    
\title{\LARGE \bf
Improving Air Mobility for Pre-Disaster Planning with Neural Network Accelerated Genetic Algorithm
}

\author{Kamal Acharya$^{1}$ Alvaro Velasquez$^{2}$ Yongxin Liu$^{3}$ Dahai Liu$^{3}$ Liang Sun$^{4}$ and Houbing Song$^{1}$
\thanks{$^{1}$K. Acharya and H. Song are with the Security and
Optimization for Networked Globe Laboratory (SONG Lab), Department of Information Systems, University of Maryland, Baltimore County, Baltimore, MD 21250 USA  {\tt\small kamala2@umbc.edu; h.song@ieee.org}}
\thanks{$^{2}$A. Velasquez is with the Department of Computer Science, University of Colorado, Boulder, CO 80309 USA
        {\tt\small alvaro.velasquez@colorado.edu}}
\thanks{$^{3}$Y. Liu and D. Liu are with the Mathematics Department and College of Aviation, Embry-Riddle Aeronautical University, Daytona Beach, USA
{\tt\small liuy11@erau.edu;liu89b@erau.edu}}
\thanks{$^{4}$L. Sun is with the Department of Mechanical Aerospace Engineering, New Mexico State University, Las Cruces, USA
        {\tt\small lsun@nmsu.edu}}
}

\begin{document}

\maketitle
\thispagestyle{empty}
\pagestyle{empty}

\begin{abstract}

Weather disaster related emergency operations pose a great challenge to air mobility in both aircraft and airport operations, especially when the impact is gradually approaching. We propose an optimized framework for adjusting airport operational schedules for such pre-disaster scenarios. We first, aggregate operational data from multiple airports and then determine the optimal count of evacuation flights to maximize the impacted airport's outgoing capacity without impeding regular air traffic. We then propose a novel Neural Network (NN) accelerated Genetic Algorithm(GA) for evacuation planning. Our experiments show that integration yielded comparable results but with smaller computational overhead. We find that the utilization of a NN enhances the efficiency of a GA, facilitating more rapid convergence even when operating with a reduced population size. This effectiveness persists even when the model is trained on data from airports different from those under test. Data and code available at "https://github.com/lotussavy/ITSC-2024".


\end{abstract}

\section{Introduction}
Emergency situations are an unavoidable phenomenon and their impact on the aviation cannot be neglected. Yearly many people move from one place to another due to the impact of fatal emergency caused by natural disasters such as hurricanes \cite{smith2013us,9294393}. Typically, in anticipation of a natural disaster, there is a significant surge in population movement towards safer areas. This leads to a marked increase in traffic across all modes of transportation, including airways, commencing approximately one week prior to the projected date of the disaster's impact. As the demand for air travel and daily commuting intensifies, the availability of aircraft becomes increasingly limited. This situation arises due to the heightened need for transportation resources during these critical times. There is an essential requirement for integrating a meticulously calibrated and validated traffic incident response module into the process of modeling and simulating evacuation scenarios\cite{zhu2019modeling}, focusing on maintaining the normal operation routine of the evacuation airport and airspace while still increasing the outgoing capability for passengers. 

An effective air mobility evacuation plan is distinguished by its capacity to not only maintain the regular flow of air traffic but also smoothly handle the increased demand for outbound flights \cite{zhou9903250}. Nevertheless, previous research that has devised evacuation plans frequently relies on utilizing the full capacity of an airport, sometimes leading to disruptions in the ongoing air mobility operations and causing panic among the people. In many instances, there is an advance anticipation of emergency situations, providing a substantial window of opportunity to develop thorough evacuation strategies prior to the occurrence of the actual emergency which has not been utilized in the previous works.

This study focuses on the efficient evacuation without disrupting the regular airport schedule. To achieve this, we make use of the non-critical capacities of the chosen airport, which are typically allocated for military (MIL) and General Aviation (GAV) operations. Through a temporary reallocation of management resources from these less critical functions, we facilitate the coexistence of evacuation flights alongside standard air traffic. The primary contributions of this study encompass:

\begin{itemize} 
    \item We propose a novel pre-disaster scheduling framework that optimize the outbound capacity of an airport affected by the ongoing emergency situation, while ensuring normal airspace operations.

    \item We explore the possibility of integrating the Genetic algorithm (GA) with a Neural Network (NN) to reduce the required number of iterations and size of population pool towards optimal solution.
    \item We investigate the generalization ability of the NN to airports on which it was not originally trained for aiding the GA.
    \item We evaluate our solution extensive with real flight operation data.

\end{itemize} 

The remainders of this paper are organized in the following manner: A review of prior research is presented in the related work, followed by the methodology. Subsequently, an evaluation and discussion of our findings are detailed, leading to the final conclusions.

\section{Related Work}
Machine learning algorithms, such as deep learning and reinforcement learning, are being increasingly explored and applied in the context of air mobility for emergency evacuation situations. These advanced computational techniques offer the potential to optimize and enhance the efficiency of air transportation during critical evacuation scenarios.

In \cite{rahman2018short}, a method to predict the time mean speed of freeways during hurricane evacuation using Long-Short Term Memory (LSTM) is presented. In \cite{zhang2021learning} a reinforcement learning based approach is proposed for emergency resource allocation in the air transportation system, specifically for hurricane evacuation. They formulated the flight dispatch as an online maximum weight matching problem, aiming to add more flights for evacuation while minimizing airspace complexity and air traffic controller workload. 

A machine learning enabled Adaptive Air Traffic Recommendation System for Disaster Evacuation is proposed in \cite{yang2021machine}, which aims to optimize the use of available resources to transport people from evacuated areas to safe places during extreme weather conditions. The system was composed of an offline learning component and an online planning component and balances the trade-off between airspace complexity and evacuation efficiency. To tackle the uncertainty in the air route network, a machine learning model was developed to predict the delay situation using real-time airport and weather information. 

Another research work in \cite{jiang2021spatial} utilized a spatial-temporal graph data mining approach for predicting air mobility. The proposed approach formulates the air transportation network as a graph structure and constructed a graph data set from airline on-time performance data. Spatial-temporal graph NNs were then applied to predict three measurements related to air mobility: number, average delay, and average taxiing time of departure and arrival flights at various airports in the United States.Several other works have delved into the utilization of machine learning models regarding historical flight data, automatic-dependent surveillance-broadcast (ADS-B) data, and meteorological data. These endeavors aimed to forecast micro-level phenomena, like flight punctuality\cite{gui2019flight,ayhan2018predicting,jiang2020applying}, and macro-level dynamics encompassing regional airspace and airport aspects, such as flight delays and traffic flow\cite{liu2019using,yang2020airport,zhang2020tree}. 

 GA has been used in few of the researches focused on the aviation scheduling. In\cite{lee2007multi}, GA was used in flight scheduling in the simulation environment  focusing on optimizing flight timings, gate assignments, and crew schedules to enhance operational efficiency and reduce costs. Another study \cite{sun2020research} used the GA to optimize the routing of cargo flights, taking into account factors such as fuel consumption, flight times, and cargo capacity.

The existing research focuses on harnessing machine learning algorithms to enhance evacuation plans and resource allocation during emergencies. Our research more specifically exploits the characteristics of approaching natural disaster scenarios, such as hurricanes, in which provide a valuable evacuation and planning window exists before the impact. During this lead-up period, we proactively optimize air traffic by scheduling additional evacuation flights in advance. This scheduling is facilitated by tapping into the non-critical aviation capabilities of airports, ensuring that a predetermined number of flights are scheduled without disrupting regular air traffic. Consequently, when the actual emergency occurs, there is no rush for the remaining evacuations. Our innovative approach not only maximizes the outbound capacity of affected airports but also places strong emphasis on safeguarding uninterrupted airspace operations.

\section{Methodology}

\subsection{Problem Formulation}

We performed a comprehensive analysis of aircraft capabilities at nine major airports in Florida. This analysis encompassed an examination of the operational records of various aircraft categories, including Air Carrier (AC), Air Taxi (AT), General Aviation (GAV), and Military (MIL) aircraft. In this context, an Air Carrier (AC) refers to an aircraft with a seating capacity exceeding 60 seats, while an Air Taxi (AT) aircraft can accommodate a maximum of 60 seats. General Aviation (GAV) covers all civilian aircraft movements involving takeoffs and landings, excluding those categorized as AC or AT aircraft. MIL aircraft activities, encompasses military takeoffs and landings as provided by the Federal Aviation Administration (FAA) and the Federal Test Center (FTC).\footnote{FAA Operations Network (OPSNET): \url{https://aspm.faa.gov/aspmhelp/index/OPSNET_Reports__Definitions_of_Variables.html }}.

For increasing airport capability before the day-of-impact, we employ the capabilities of General Aviation and Military Operations to minimize the impact on regular Air Carrier and Air Taxi traffic. We want to minimize the potential delay of these newly added flight while keep them favorable to passengers. To achieve this objective, we introduced two metrics: the mean of combined non-commercial capability denoted as $c$ and its standard deviation represented by $s$. $c$ signifies the average value of the number of operations (landing and takeoff) of GAV and MIL aircraft at the airports calculated from historical data. This metric serves as a reference for evaluating the effectiveness of airport capabilities during emergency scenarios. For each destination airport, the two metrics at the time-of-arrival gauge its potential capability to accept incoming flights. 

Mathematically, our objective is to devise an hourly evacuation flight schedule that maximizes $\Sigma c$ and simultaneously minimizes $\Sigma s$ across all added flights. And thus optimizing the use of GAV and MIL capabilities while ensuring efficient and reliable evacuation procedures.

\begin{figure}[htbp]
\centerline{\includegraphics[width=\columnwidth]{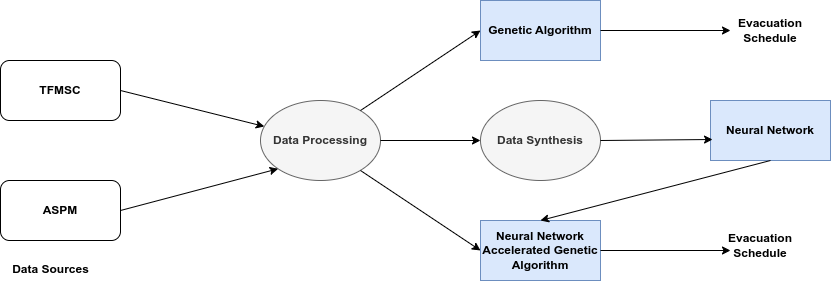}}
\caption{Research Process}
\label{fig:researchMethodology}
\end{figure}

The complete research framework of our project is presented in \autoref{fig:researchMethodology}, where our primary focus was the utilization of GA to formulate an evacuation plan. We endeavored to enhance the efficiency of this process through the incorporation of NN.

\subsection{Data Processing}
Our dataset is gathered from the FAA data repository for the initial two months of 2023. The data was sourced from multiple datasets to comprehensively address our research objectives, which encompass the following:
\begin{itemize}
    \item Traffic Flow Management System Counts (TFMSC)\footnote{FAA Traffic Flow Management System Counts (TFMSC): \url{ https://aspm.faa.gov/tfms/sys/OPSNET.asp }}:  This dataset aids in assessing the combined capabilities of the airports which are essential for gauging the potential of GAV and MIL operations during evacuation scenarios.
    \item Aviation System Performance Metrics (ASPM)\footnote{FAA Aviation System Performance Metrics (ASPM): \url{ https://aspm.faa.gov/apm/sys/AnalysisCP.asp }}: The ASPM dataset is accessible through an online access system provided by the FAA and delivers comprehensive data regarding flights to and from the ASPM airports, as well as all flights operated by ASPM carriers. We can use its data for city pair analysis, determining flight duration and retrieving the top ten destinations airports.
\end{itemize}

\begin{figure}[htbp]
\centerline{\includegraphics[width=\columnwidth]{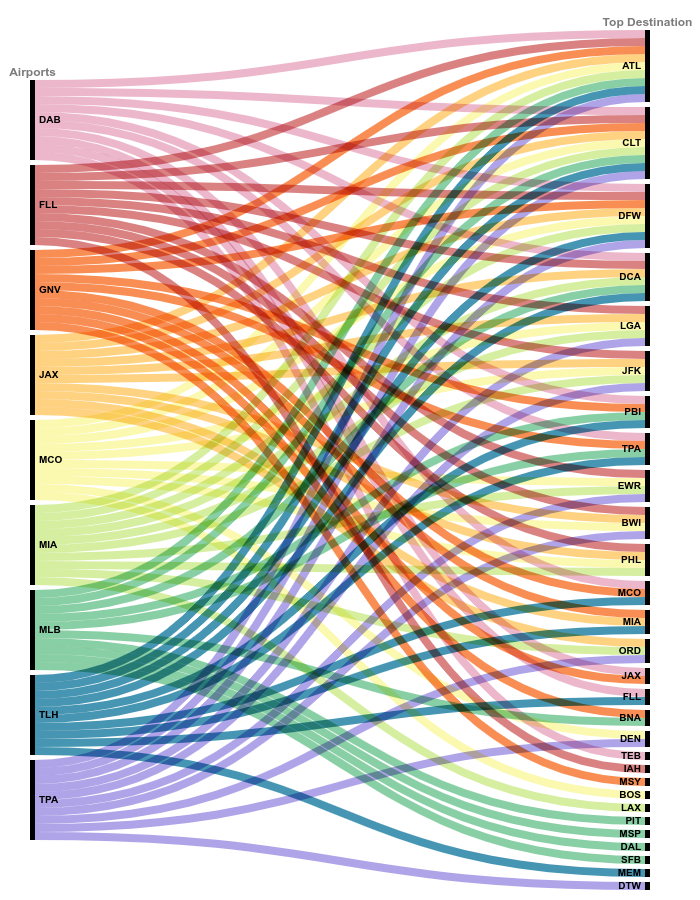}}
\caption{Top ten destination airports from the nine major airports in Florida}
\label{fig:topdestinations}
\end{figure}

We structured the TFMSC dataset on an hourly basis, and for each hourly time slot, we calculated two significant statistical metrics: the mean of combined capability $c$ and its corresponding standard deviation $s$. We combined data from 9 major Florida airports and their corresponding top ten destination outside Florida as in \autoref{fig:topdestinations}.

\subsection{Genetic Algorithm}

The GA operates by identifying optimal chromosomes from a population pool, employing the principles of crossover and mutation over multiple generations. In the context of our specific problem, a chromosome is symbolized as a list containing 10 elements, with each element assigned a value of either 1 or 0. These 10 elements correspond to the decision of whether to select or reject flights to the top ten destination airports from the evacuating airport. The effectiveness of a chromosome is evaluated through a fitness function. This fitness function is formulated using the equation specified below:
\begin{equation}
  FitnessScore= 0.5 * p + 0.2* c - 0.3 * s - penalty
\end{equation}
where
\begin{itemize}
    \item $p$ is the popularity value of the destination airport based on the number of flights from evacuating airport
    \item $c$ is the mean of the combined capability (defined in Section III-A)
    \item $s$ is the standard deviation of the combined capability (defined in Section III-A)
    \item $penalty$ equals to 1 when choosing more number of flights in destination airports beyond the capability of evacuating airport
\end{itemize}

To enhance the effectiveness of our GA, several strategies were incorporated to optimize the evacuation strategy selection process. These strategies were fine-tuned to prioritize certain factors and balance trade-offs during decision-making.
\begin{itemize}
\item \textit{Positive Influence of Popularity and Capability}:
In the pursuit of creating more effective evacuation strategies, the algorithm was configured to place a positive emphasis on $p$ and $c$. This incentives the algorithm to prioritize airports with higher popularity and greater capabilities in order to to maximize the utilization of well-equipped and frequently used airports.
\item \textit{Minimization of Delay Expectations}:
To mitigate potential delays arising from disparities in the $c$ across different hours of the day, $s$ was introduced with a negative weight which prompts the algorithm to favor airports with lower standard deviation values, thus reducing the variability in evacuation efficiency.
\item \textit{Penalties for Over-complicated Planning}:
In order to avoid an unrealistic scenario where the algorithm selects too many airports, a penalty mechanism was introduced which comes into play when the selection of airports surpasses the combined capability of the evacuating airport. By penalizing such instances, the algorithm is encouraged to strike a balance between maximizing the fitness score and adhering to practical constraints.
\end{itemize}
The complete GA implementation, encompassing these strategies, is presented in \hyperref[alg:GeneticAlgorithm]{Algorithm 1}.

\begin{algorithm}[tb]
\caption{GeneticAlgorithm}
\label{alg:GeneticAlgorithm}
\textbf{Input}:Population size($population\_size$), number of generations($num\_generation$), crossover rate, mutation  rate,  mean of combined capability of evacuating airport(c), dataset(d) containing details of destination airports\\
\textbf{Output}: The best selection list found in the last generation.
\begin{algorithmic}[1] 
\STATE Initialize the population of ($population\_size$)
\STATE Evaluate the fitness score of each individual in population
\STATE Let $gen=1$
\WHILE{$gen<num\_generation$}
\STATE Create an empty new population
\WHILE{$size(new population) < population\_size$}
\STATE Select two parents, parent1 and parent2, from population based on their fitness score(higher fitness score has higher chance of selection)
\STATE Perform crossover operation by selecting the random crossover point to create two children, child1 and child2, from parent1 and parent2
\STATE Perform mutation operation on child1 and child2 by iterating bit by bit and randomly flipping the bit whenever the random number generated is less than the mutation rate
\STATE Add child1 and child2 to new population
\ENDWHILE
\STATE Replace entire old population with new population
\STATE Evaluate the fitness of each individual in new population
\ENDWHILE
\STATE best selection list $\leftarrow$ individual in population with fitness equal to best fitness score
\STATE \textbf{return} best selection list
\end{algorithmic}
\end{algorithm}

\subsection{Data Synthesis}
For the purpose of training the neural network, an extra step involving data synthesis was introduced. In this process, the airport for which the evacuation strategy was under development was deliberately omitted from the dataset. This was done to ensure that the model did not have access to the testing data while being trained. In our case, we removed data of DAB airport to synthesize the training data for NN. Data from all other eight airports was amalgamated, encompassing attributes such as the mean combined capability ($c$), and the standard deviation ($s$) of $c$ along with the popularity ($p$) of the top ten destination airport and their corresponding $c$ and $s$. From these attributes the best possible combination of the destination airports were determined and added to complete the dataset. Overall algorithm for synthesizing the data frame is given in \hyperref[alg:dataGeneration]{Algorithm 2}.

\begin{algorithm}[tb]
\caption{Algorithm for generating dataset for Neural Network}
\label{alg:dataGeneration}
\textbf{Input}:Historical flight data from  TFMSC and ASPM having N entries\\
\textbf{Output}: Synthesized dataset containing containing 41 columns, 30(A1 to A30) for each $p$,$c$ and $s$ values of top ten destination representing popularity, mean capability and its standard deviation respectively, 1(C) for mean Capability of evacuating airport and 10(S1 to S10) for representing selection and rejection of the top ten destination.
\begin{algorithmic}[1] 
\STATE Let $i=0$.
\WHILE{$i<N$}
\STATE Create all possible 1024 combination of selection list $[S1,S2,S3,S4,S5,S6,S7,S8,S9,S10]$
\STATE Calculate fitness score for each of them based on equation $1$
\STATE Let $j=0, bestFitness=0,bestCombination=[]$
\WHILE{$j<1024$}
\IF {$fitness>bestFitness$}
\STATE bestFitness = fitness, bestCombination = combination
\ENDIF
\ENDWHILE
\STATE ADD mean Capability of evacuating airport and bestCombination to the corresponding row of the dataset
\ENDWHILE
\STATE \textbf{return} synthesized dataset
\end{algorithmic}
\end{algorithm}

\subsection{Neural Network Predictor for Optimal Solution}

\begin{figure}[htbp]
\centerline
{\includegraphics[width=\columnwidth]{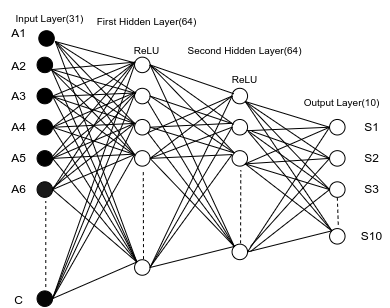}}
\caption{Neural Network Architecture}
\label{fig:NNArchitecture}
\end{figure}
A simple NN model consisting of two hidden layers, as depicted in \autoref{fig:NNArchitecture}, was deployed to enhance the efficiency of GA. This model underwent training using a synthetic dataset generated by \hyperref[alg:dataGeneration]{Algorithm 2}. The NN model takes in 31 input variables, marked A1 to A30, that capture the popularity, mean capability, and the standard deviation in mean capability of the top ten destination airports for evacuation. In addition to these, there is an additional input, designated as $C$, which specifies the mean capability of the evacuating airport. The model's output is a 10-bit representation identifying the chosen destination airports for evacuation. Through training with the synthetic dataset, the model was able to learn patterns and connections among various attributes, thereby gaining the ability to accurately predict which destination airports should be selected for evacuation at any given hour of the day.

\subsection{Integration of Neural Network to Genetic Algorithm}
In this phase, we integrated the GA given in \hyperref[alg:dataGeneration]{Algorithm 2} with the trained NN model to facilitate a quicker convergence of the algorithm. We aim to harness the predictive power of the NN to initialize the population for the GA. In line 12 of \hyperref[alg:dataGeneration]{Algorithm 2} instead of replacing the entire old population with the new population obtained by GA, we introduced NN for also contributing to the population pool. This process of generating the population by NN and adding to the population tool involved two approaches:

\textbf{Approach 1: }Population generated by the NN are directly added to the pool by replacing the random population generated by the GA.

\textbf{Approach 2: } In this approach before inserting the population generated by NN, the population is sorted in descending order on the basis of fitness score. The lower order of population are replaced by the population generated by NN.

\section{Evaluation and Discussion}
This section focuses in evaluating the performance, accuracy, and efficiency of the developed methodologies. 

\subsection{Genetic Algorithm}
\begin{figure}[htbp]
\centerline
{\includegraphics[width=\columnwidth]{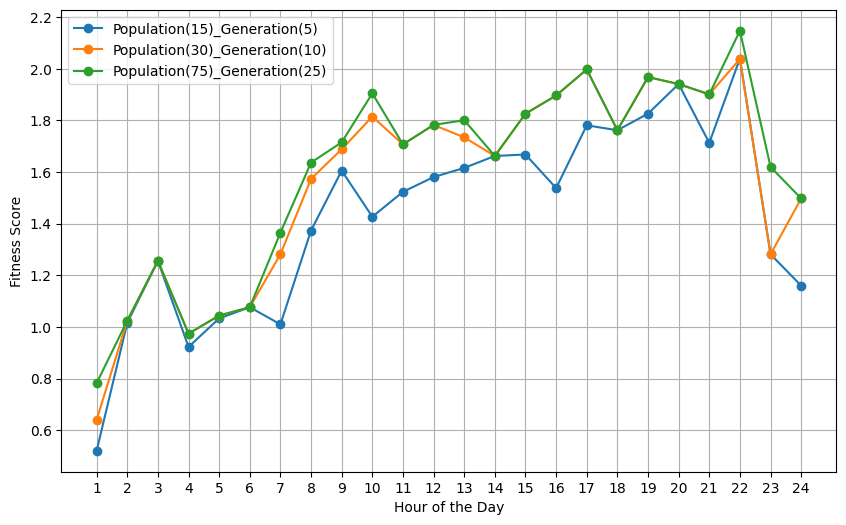}}
\caption{Fitness Score for various combination of population size and number of generation for Genetic Algorithm during different hour of the day. }
\label{fig:GA}
\end{figure}

We used only GA to find the optimal evacuation flight schedules originating from DAB airport. The algorithm was initiated with a population size of 15 and executed for 5 generations. The population size and number of generations were systematically increased by factors of two and five to explore their impact on the algorithm's performance. During this iterative process, it was observed that the algorithm exhibited a substantial convergence trend when the population size was set to 75 and the number of generations was increased to 25, as in \autoref{fig:GA}. The observed convergence trend suggests that a population size of 75 with 25 generations struck a balance between exploration and exploitation, leading to optimal results in terms of identifying the most effective evacuation flight schedules from the selected airport.

\subsection{Neural Network}
\begin{figure}[htbp]
\centerline{
\includegraphics[width=\columnwidth]{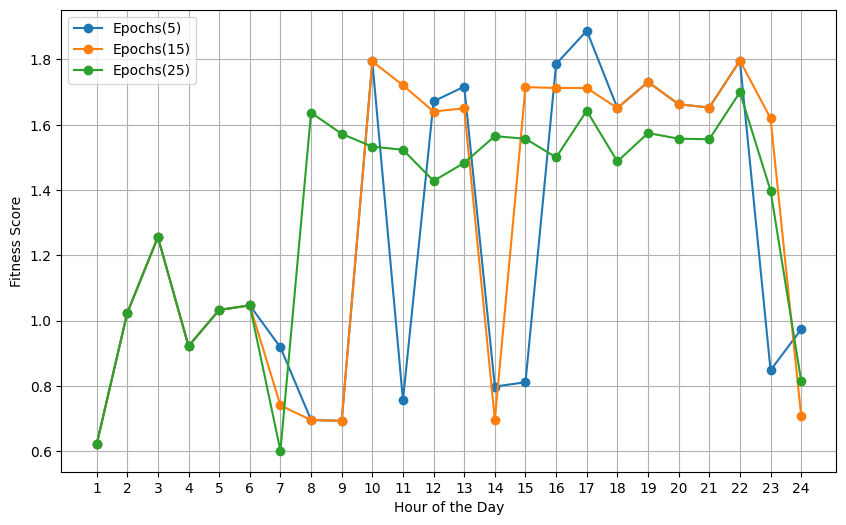}} 
\caption{Fitness Score for various neural networks trained under different number epochs}
\label{fig:NN}
\end{figure}

The NN was trained using the synthesized dataset outlined in \hyperref[alg:dataGeneration]{Algorithm 2}. During training, the NN was exposed to varying numbers of epochs, namely, 5, 15, and 25. And it was observed that convergence commenced at approximately 25 epochs. Prior to convergence, there was a notable performance fluctuation as depicted in \autoref{fig:NN}. 

\subsection{Combination of Genetic Algorithm and Neural Network}
To demonstrate the effectiveness of the Neural Network-accelerated Genetic algorithm. We select sub-optimal scenarios for each method individually and merge them to find an optimal scenario. Specifically, the GA with a population size of 15 and 5 generations and the NN trained for only 5 epochs respectively. The combined approach, referred to as the "NN-accelerated GA," leveraged the predictive capabilities of the NN to enhance the parent population generation for the GA. The strategy entailed using the NN to produce parent candidates, which were then integrated into the parent pool. The GA was then employed to refine the parent pool further. Parents were added based on the fitness score, considering the NN-generated parents alongside those generated by the GA itself.

\begin{figure}[htbp]
\centerline{
\includegraphics[width=\columnwidth]{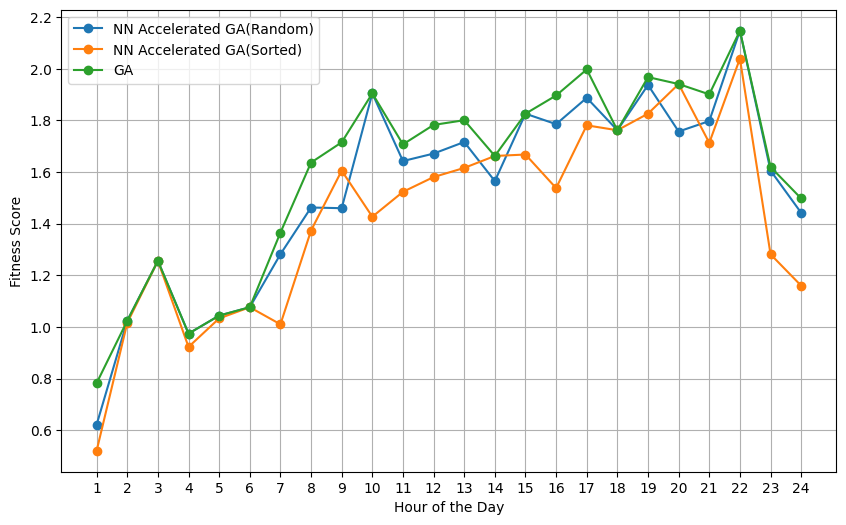}}
\caption{Fitness Score of GA and two differnt NN accelerated GA}
\label{fig:types}
\end{figure}

\begin{figure}[htbp]
\centerline{
\includegraphics[width=\linewidth]{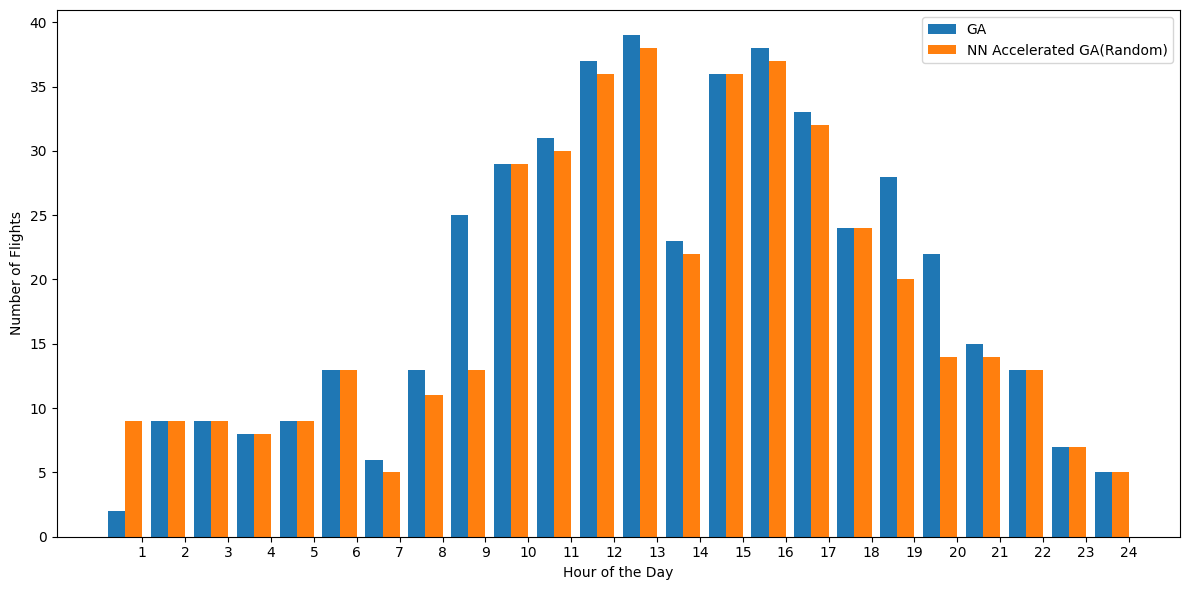}} 
\caption{Number of evacuation flights scheduled by the GA (having population size of 75, iterated for 25 generations) and NN acclerated GA(NN trained for 5 epochs and randomly replacing the population in population pool of GA having population size of 15 and iterated for 5 generations). }
\label{fig:flights}
\end{figure}

\begin{figure}[htbp]
\centerline{
\includegraphics[width=\linewidth]{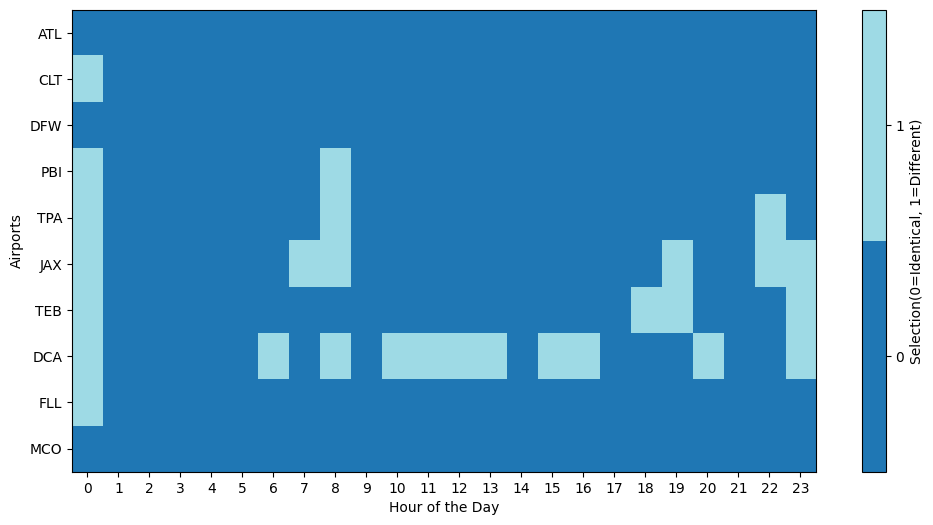}} 
\caption{Comparison of selection of the airports for evacuation by GA(having population size of 75 and iterated for 25 generations) and NN accelerated GA(NN trained for 5 epochs and randomly replacing the population in population pool of the GA having population size of 15 and iterated for 5 generations )}
\label{fig:selection}
\end{figure}
We compared two methods: random addition of NN-generated parents and the selective removal of the least-fit population. In the latter approach, we ranked the entire population pool in descending order based on their fitness scores. The least-fit populations were then replaced by the parents generated by the NN. The results from this experiment, as depicted in \autoref{fig:types}, revealed that the strategy of random addition of parents from the NN yields better outcomes than the approach of selectively removing the least-fit populations.

As depicted by \autoref{fig:flights}, we found that the prediction of the number of flights by GA iterated for 25 generations with population size of 75 and NN accelerated GA with the strategy of random removal , is almost identical except in the midnight scenario. The reason might be due to the fact the number of flights in the DAB airport in that time duration is very random sometimes they have few flights but most of the time number of flights during those time is almost none. Another \autoref{fig:selection} provided detail about the selection of top ten destination airports by same two models. As shown in the figure we found that most of the time airport selection by both of them are identical.

\section{Conclusion}
In this research, we delved into the utilization of GA to develop a novel pre-disaster evacuation planning framework for advanced air mobility. We leverage the non-commercial flight capability of impacted airports to ensure that evacuation plans minimally affected routine airspace operation. We proposed a neural network accelerated genetic algorithm to derive our solution with smaller computational overhead. We found that GA exhibited slower convergence when operated independently with higher population pool and generations, its performance significantly improved when assisted with a NN model trained for a mere 5 epochs. This integration yielded comparable results in terms of fitness function, flight numbers, and airport selection. This intriguing finding underscores the NN's ability to expedite GA's convergence, even when trained on data from different airports, showcasing its generalization capabilities.

As a prospect for future exploration, we are planning to incorporate another NN to predict fitness scores. This endeavor aims to ascertain whether the fusion of these two NN models could yield even faster convergence rates when combined with our current hybrid model.


\section*{ACKNOWLEDGMENT}
This research was supported by the Center for Advanced Transportation Mobility (CATM), USDOT Grant
\#69A3551747125.


\bibliographystyle{IEEEtran}
\bibliography{ref.bib}

\end{document}